\documentclass[copyright,creativecommons]{eptcs}
\providecommand{\event}{FMAS 2026}

\usepackage{iftex}
\ifpdf
  \usepackage{underscore}
  \usepackage[T1]{fontenc}
\else
  \usepackage{breakurl}
\fi

\usepackage{xspace}
\usepackage{url}
\usepackage{graphicx}
\usepackage{xcolor}
\usepackage{booktabs}
\usepackage{tabularx}
\usepackage{float}
\usepackage{array}
\usepackage{placeins}
\usepackage{listings}
\usepackage{tikz}
\usetikzlibrary{arrows.meta,backgrounds,calc,fit,positioning}

\newcommand{\ros}{ROS\xspace}
\newcommand{\rostwo}{ROS~2\xspace}
\newcommand{\rosmon}{ROSMonitoring\xspace}
\newcommand{\code}[1]{\texttt{#1}}

\definecolor{listinggray}{gray}{0.25}
\lstdefinelanguage{rosyaml}{
  morekeywords={monitors,monitor,id,log,oracle,url,port,actions,name,type,action,clients,servers},
  sensitive=true,
  morecomment=[l]{\#},
  morestring=[b]",
  morestring=[b]'
}
\title{ROSMonitoring in Action: Runtime Monitoring and Enforcement for ROS 2 Actions}

\author{Giacomo Paltrinieri
\institute{ Department of Physics, Informatics and Mathematics\\
University of Modena and Reggio Emilia, Italy}
\email{314875@studenti.unimore.it}
\and 
Angelo Ferrando
\institute{ Department of Physics, Informatics and Mathematics\\
University of Modena and Reggio Emilia, Italy}
\email{angelo.ferrando@unimore.it}
}
\def\titlerunning{\rosmon for \rostwo Actions}
\def\authorrunning{G. Paltrinieri \& A. Ferrando}

\begin{document}
\maketitle

\begin{abstract}
Robotic applications often rely on long-running operations, such as navigation, manipulation, and inspection, which in ROS~2 are commonly implemented as actions. Unlike topics and services, actions involve a structured lifecycle comprising goal submission and acceptance, feedback, status updates, result retrieval, and cancellation. Existing versions of ROSMonitoring support topics and services, but do not reconstruct this lifecycle or expose action-level events to a runtime-verification oracle.
In this paper, we extend ROSMonitoring with runtime monitoring and enforcement for ROS~2 actions. Actions are declared directly in the monitoring configuration and automatically instrumented through the underlying ROS~2 communication endpoints. The generated monitor reconstructs action-level events, correlates them through goal identifiers, maintains the state of active goals, and translates violations into protocol-aware enforcement operations. These include goal rejection, cancellation, message suppression, and protocol-compliant responses generated directly by the monitor when the original communication cannot be forwarded.
We evaluate the implementation through controlled stress tests that vary the number of action invocations, workload size, and client--server topology. The results show approximately linear growth in monitoring cost with repeated action invocations under a fixed workload, while larger action workloads and a greater number of configured client--server pairs introduce additional overhead. These results characterise the scaling behaviour of the current implementation under the tested sequential workloads while highlighting the main performance limitations of the current synchronous implementation.
\end{abstract}

\section{Introduction}
\label{sec:introduction}

There are many techniques for checking the reliability of software systems.
Some analyse a model before the system is deployed.  Others observe the system
while it is running.  Runtime verification (RV) belongs to the second
group~\cite{LeuckerSchallhart2009}.  An RV monitor receives events from one
execution and checks whether they satisfy a formal property.  The property can
be checked online, as events are produced, or offline, over a recorded trace.

Runtime verification is useful for robotic applications because robots
interact with changing environments.  Not every execution can be predicted at
design time.  It is therefore useful to observe the actual execution and to
react when a required property is violated.

The Robot Operating System (\ros) provides a common software platform for
robotics.  \rostwo is used in many research and industrial systems and
supports topics, services, and actions~\cite{MacenskiEtAl2022}.  Topics carry
streams of messages, while services implement request--response communication.
Actions are used for operations that may take time to complete.  A client
sends a goal to an action server and receives an acceptance decision.  While
the goal is running, the server can publish feedback and status updates.  The
client can later request the result or ask the server to cancel the goal.

Although an action is presented to the developer as one communication
mechanism, \rostwo implements it using two topics and three services.  These
endpoints carry different parts of the same goal lifecycle and relate to one
another through goal identifiers.  Monitoring them as unrelated topics and
services loses this action-level information.

\rosmon was introduced as a formalism-agnostic RV framework for
\ros~\cite{FerrandoEtAl2020}.  Its main idea is to separate the instrumentation
of the robotic application from the oracle that implements the chosen
specification language. Previous versions of \rosmon support topic and service
monitoring~\cite{FerrandoEtAl2020,GhaffariSaadatEtAl2024},
but do not reconstruct the lifecycle of \rostwo actions.

An action monitor needs to know which lifecycle event is being observed and
which goal it belongs to.  It must remember accepted goals so that later
feedback, status, result, and cancellation events can be correlated with the
correct request. Enforcement must preserve the ROS 2 action protocol. When a goal request violates a property, the monitor cannot simply discard it, because the client would remain waiting for a response. Instead, the monitor must return a valid \code{SendGoal.Response} with \code{accepted=false}. Similarly, if a violation is detected while processing a feedback event, suppressing that feedback only prevents the client from observing it; it does not stop the goal already being executed by the server.
In this paper, we extend \rosmon with monitoring and enforcement support for
\rostwo actions.  The extension keeps the existing topic and service
monitoring infrastructure, but generates it from a single action declaration
and adds the missing action semantics.  In particular, the extension provides:
\begin{itemize}
  \item action-level instrumentation that expands one configured action into
        the two topics and three services used by the \rostwo action protocol;
  \item action events enriched with the lifecycle kind, action name, and goal
        identifier, together with state for accepted goals;
  \item correlation of requests, responses, feedback, status updates, results,
        and cancellations belonging to the same action goal;
  \item protocol-aware enforcement that can reject a new goal, cancel an
        accepted goal, suppress violating feedback or status updates, and
        construct valid rejection or aborted-result responses; and
  \item an evaluation combining a robotic case study with a controlled
        scalability benchmark.
\end{itemize}

The action state maintained by the extension allows ROSMonitoring to operate
as an external coordination layer in systems composed of multiple independent
\rostwo frameworks. By mediating action communication, it can enforce
system-level properties that depend on the combined state of those frameworks. We demonstrate this through a simulated
mobile manipulator that combines Nav2, MoveIt~2, and
\code{ros2_control}.  The monitor enforces an external safety contract that
allows navigation only when the arm is in a safe configuration and rejects
new arm-trajectory or gripper-opening goals while navigation is active. The case study
shows that compliant goals are forwarded normally, while conflicting goals
are rejected without modifying Nav2, MoveIt~2, or the robot controllers.

We complement this functional evaluation with a synthetic scalability
benchmark.  The benchmark varies the number of configured client--server
pairs, the number of goals sent by each pair, and the amount of data returned
by each action.  Every configuration is executed five times in both baseline
and monitored conditions, and the complete experimental configuration is
restarted for every repetition.

The quantitative results show that, for one fixed client--server pair and a
fixed result length, the added time is close to proportional to the number of
action calls.  However, the cost cannot be described by one universal
per-action value.  Longer results increase the number or size of monitored
events, and configurations with more client--server pairs introduce an
additional topology effect.  The evaluation therefore demonstrates the functional usefulness of
action-aware enforcement and characterises the overhead of the current
synchronous implementation.

The rest of the paper is organised as follows.
Section~\ref{sec:background} recalls the \rostwo action protocol and the
architecture of \rosmon.
Section~\ref{sec:framework} presents the action declaration, generated
endpoints, event correlation, goal state, and protocol-aware enforcement.
Section~\ref{sec:evaluation} presents the qualitative case study and the
quantitative scalability experiments.
Section~\ref{sec:related-work} discusses related work on runtime verification
for robotic systems and \ros.
Finally, Section~\ref{sec:conclusion} concludes the paper.

\section{Background}
\label{sec:background}

This section recalls the parts of \rostwo and \rosmon needed to present the
extension.  We first describe how actions are implemented in \rostwo.  We then
summarise the monitoring architecture provided by the previous versions of
\rosmon.

\subsection{\rostwo Actions}
\label{sec:background-actions}

\rostwo provides topics, services, and actions as its main communication
mechanisms~\cite{MacenskiEtAl2022}.  A topic supports asynchronous
communication between publishers and subscribers.  A service supports a
request--response interaction between a client and a server.  An action
supports a longer-running interaction in which a client submits a goal and
receives information about its execution over time.

An action interaction starts when the client sends a goal request.  The action
server can accept or reject the goal.  If the goal is accepted, the server can
publish feedback while executing it.  The client can request the final result
or ask the server to cancel the goal.  The server also publishes status
information about active and completed goals.  An accepted goal may eventually
succeed, be aborted, or be cancelled.

Each goal is identified by a UUID.  This identifier connects the initial goal
request with later feedback, status, result, and cancellation messages.
Several goals may be active at the same time, including different instances of
the same action.  Action-aware monitoring must therefore preserve the
association between each event and its goal.

Although an action is exposed to the developer as one communication
mechanism, it is implemented using three services and two
topics~\cite{ROS2ActionsDesign}.  For an action named \code{<name>}, these
endpoints are:
\begin{itemize}
  \item the \code{<name>/_action/send_goal} service, used to submit a goal and
        return the acceptance decision;
  \item the \code{<name>/_action/get_result} service, used to request and
        return the final result;
  \item the \code{<name>/_action/cancel_goal} service, used to request the
        cancellation of one or more goals;
  \item the \code{<name>/_action/feedback} topic, used to publish
        goal-specific progress information; and
  \item the \code{<name>/_action/status} topic, used to publish the state of
        active and recently completed goals.
\end{itemize}

These endpoints form one protocol.  Their messages cannot always be monitored
or modified independently.  For example, suppressing the response that accepts
a goal does not remove the goal from the server.  Once the server has accepted
it, stopping the operation requires a cancellation request containing the
corresponding UUID.  Similarly, suppressing feedback prevents the client from
seeing an update but does not stop the action itself.

\subsection{\rosmon}
\label{sec:background-rosmonitoring}

\rosmon is a formalism-agnostic framework for runtime monitoring of robotic
applications~\cite{FerrandoEtAl2020}. Its architecture comprises a generator,
generated ROS-facing monitors, and a separate verification oracle. These
components separate application instrumentation from property evaluation.

The generator reads a YAML configuration identifying the application nodes,
the communication channels to intercept, and whether their traffic should be
logged or filtered. It produces the monitoring nodes and the remappings
required to route the selected communication through them. At runtime, the
generated monitors intercept messages and convert them into JSON events.

In logging mode, events are stored for offline analysis. In online filtering
mode, the monitor sends each event to the oracle over a WebSocket connection
and waits for a verdict before deciding whether to forward the intercepted
communication. The oracle implements the selected property language and
verification procedure; the ROS-facing monitor applies the resulting
communication decisions. This interface allows different oracle
implementations and specification formalisms to be used without changing
the application instrumentation.
A topic proxy can observe, record, and
filter published messages.  A service proxy can check both the request and the
response while preserving the request--response interaction.

These mechanisms provide most of the transport-level support required for
actions because an action is built from topics and services.  However,
configuring the five action endpoints separately does not recover their
shared semantics.  The generated events do not identify the action-lifecycle
step, and the monitor does not automatically correlate messages belonging to
the same goal.  The existing filtering rules also do not determine which
protocol-level response is required when a violation occurs.

The extension presented in the next section builds on the existing ROSMonitoring architecture while adding action-aware generation, event correlation, state management, and enforcement.

\section{\rosmon with Actions}
\label{sec:framework}

As discussed in Section~\ref{sec:background}, a \rostwo action is implemented
through two topics and three services.  \rosmon 2.0 can already generate
proxies for these endpoints.  However, five independent proxies do not tell
the oracle which action step is being observed or which goal is involved.
They also do not determine how a violating event should be handled without
breaking the action protocol.

The extension adds this missing action-level information while keeping the
existing \rosmon architecture. The application\footnote{\url{https://github.com/GiacomoPaltrinieri/ROSMonitoring/tree/ros2}} still uses the standard
\rostwo \code{ActionClient} and \code{ActionServer} interfaces, and the
oracle still exchanges JSON events and verdicts with the monitor.  The
extension changes two parts of the framework.  First, the generator compiles
one action declaration into the five topic and service proxies used by
\rostwo.  Second, the generated callbacks recognise action traffic, enrich
it with lifecycle and goal information, maintain state for accepted goals,
and apply action-specific responses to violations.

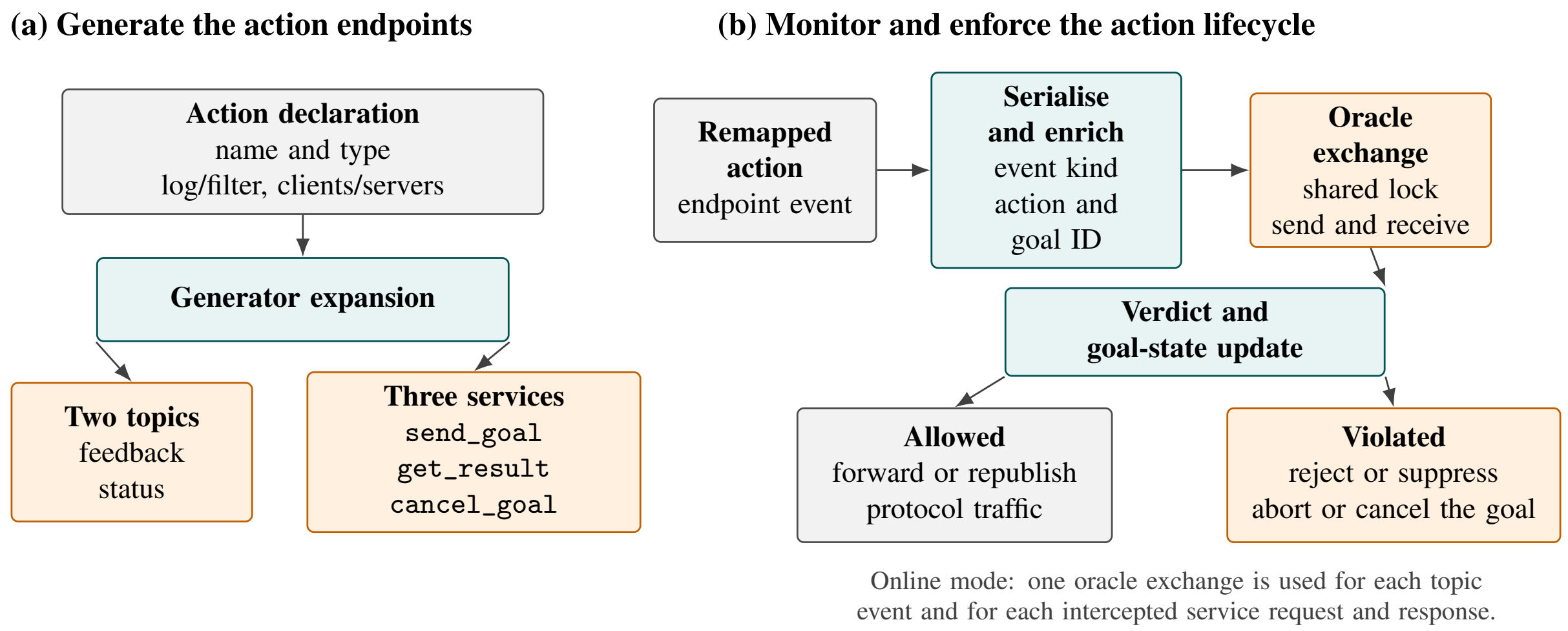
\begin{figure}[t]
  \centering
  \scalebox{1.0}{
%
\resizebox{\linewidth}{!}{%
\begin{tikzpicture}[
  font=\footnotesize,
  >={Latex[length=2mm]},
  flow/.style={->,draw=black!75,line width=0.55pt},
  basebox/.style={
    draw=black!70,
    rounded corners=2pt,
    align=center,
    line width=0.55pt,
    inner xsep=5pt,
    inner ysep=4pt
  },
  existing/.style={basebox,fill=black!5},
  generated/.style={basebox,draw=teal!65!black,fill=teal!9},
  checked/.style={basebox,draw=orange!75!black,fill=orange!12},
  paneltitle/.style={font=\small\bfseries,anchor=west},
  note/.style={font=\scriptsize,align=center,text=black!75}
]

\node[paneltitle] at (0,3.70)
  {(a) Generate the action endpoints};

\node[existing,text width=4.85cm,minimum height=1.35cm]
  (declaration) at (3.30,2.35)
  {\textbf{Action declaration}\\
   name and type\\
   log/filter, clients/servers};

\node[generated,text width=4.10cm,minimum height=0.90cm]
  (generator) at (3.30,0.75)
  {\textbf{Generator expansion}};

\node[checked,text width=2.25cm,minimum height=1.50cm]
  (topics) at (1.45,-0.90)
  {\textbf{Two topics}\\
   feedback\\
   status};

\node[checked,text width=3.25cm,minimum height=1.50cm]
  (services) at (5.15,-0.90)
  {\textbf{Three services}\\
   \texttt{send\_goal}\\
   \texttt{get\_result}\\
   \texttt{cancel\_goal}};

\draw[flow] (declaration) -- (generator);
\draw[flow] (generator.south west) -- (topics.north);
\draw[flow] (generator.south east) -- (services.north);

\node[paneltitle] at (7.65,3.70)
  {(b) Monitor and enforce the action lifecycle};

\node[existing,text width=2.05cm,minimum height=1.55cm]
  (endpoint) at (8.3,2.15)
  {\textbf{Remapped action}\\
   endpoint event};

\node[generated,text width=2.35cm,minimum height=1.55cm]
  (enrich) at (11.45,2.15)
  {\textbf{Serialise and enrich}\\
   event kind\\
   action and goal ID};

\node[checked,text width=2.25cm,minimum height=1.55cm]
  (oracle) at (14.85,2.15)
  {\textbf{Oracle}\\
   \textbf{exchange}\\
   shared lock\\
   send and receive};

\node[generated,text width=3.75cm,minimum height=0.95cm]
  (verdict) at (12.95,0.40)
  {\textbf{Verdict and}\\
   \textbf{goal-state update}};

\node[existing,text width=3.05cm,minimum height=1.45cm]
  (allowed) at (10.35,-1.15)
  {\textbf{Allowed}\\
   forward or republish\\
   protocol traffic};

\node[checked,text width=3.25cm,minimum height=1.45cm]
  (violated) at (15.10,-1.15)
  {\textbf{Violated}\\
   reject or suppress\\
   abort or cancel the goal};

\draw[flow] (endpoint) -- (enrich);
\draw[flow] (enrich) -- (oracle);
\draw[flow] (oracle.south) -- (verdict.north east);
\draw[flow] (verdict.south west) -- (allowed.north);
\draw[flow] (verdict.south east) -- (violated.north);

\node[note,text width=7.60cm] at (12.75,-2.48)
  {Online mode: one oracle exchange is used for each topic event and for each
   intercepted service request and response.};

\end{tikzpicture}%
}}
  \caption{Overview of action generation and lifecycle-aware monitoring.}
  \label{fig:architecture}
\end{figure}

Figure~\ref{fig:architecture} shows these two parts.  Endpoint interception
and oracle communication reuse the mechanisms of \rosmon 2.0.  The new code
performs action expansion at generation time and adds event enrichment, goal
correlation, state, and protocol-aware filtering at runtime.

\subsection{Action Declaration and Endpoint Generation}
\label{sec:instrumentation}

Instrumentation is configured through the same YAML file used by earlier
versions of \rosmon.  The user can add an \code{actions} list to a monitor.
Each entry gives the action name, its \rostwo type, and the operation assigned
to the generated endpoints.  The available operations are \code{log} and
\code{filter}.  The protocol-aware reactions described below concern online
filtering.

The optional \code{clients} and \code{servers} fields identify the nodes that
must be remapped.  The client declarations are applied to the three action
services.  The server declarations are applied to the feedback and status
topics.  Listing~\ref{lst:action-config} shows the main fields for a navigation
action.  The corresponding node and launch-file declarations are omitted to
keep the example short.

\begin{lstlisting}[language=rosyaml,
  caption={Main configuration fields for monitoring a navigation action.},
  label={lst:action-config}]
monitors:
  - monitor:
      id: navigation_monitor
      log: ./navigation_events.jsonl
      oracle:
        url: 127.0.0.1
        port: 8080
        action: nothing
      actions:
        - name: /navigate_to_pose
          type: nav2_msgs.action.NavigateToPose
          action: filter
          clients: [navigation_client]
          servers: [navigation_server]
\end{lstlisting}

Before the normal monitor code is generated, each action entry is expanded
into the endpoints shown in Table~\ref{tab:endpoints}.  The expansion reuses
the existing topic and service generation path.  It also converts the action
types into the internal runtime classes used by \code{rclpy}.  The status
subscriber uses the standard \rostwo action-status quality-of-service
profile.  The instrumented launch files remap the selected client and server
traffic through the generated monitor.

\begin{table}[H]
  \centering
  \caption{Endpoints generated from an action named \code{<name>}
  with type \code{<Action>}.}
  \label{tab:endpoints}
  \small
  \begin{tabularx}{\linewidth}{@{}l X@{}}
    \hline\hline
    Endpoint & Runtime type used by the generated monitor \\
    \hline\hline

    \code{<name>/_action/feedback} &
    \code{<Action>.Impl.FeedbackMessage} \\
    \hline

    \code{<name>/_action/status} &
    \code{action_msgs.msg.GoalStatusArray} \\
    \hline

    \code{<name>/_action/send_goal} &
    \code{<Action>.Impl.SendGoalService} \\
    \hline

    \code{<name>/_action/get_result} &
    \code{<Action>.Impl.GetResultService} \\
    \hline

    \code{<name>/_action/cancel_goal} &
    \code{action_msgs.srv.CancelGoal} \\
    \hline\hline
  \end{tabularx}
\end{table}

The user therefore describes a complete action once instead of listing five
internal endpoints.  The original \code{topics} and \code{services} fields
remain available when a deployment needs to observe or filter only selected
action endpoints.  In both cases, action semantics are recognised from the
standard \code{/_action/} endpoint names.

\subsection{Lifecycle Events and Goal State}
\label{sec:action-events}

When a generated callback receives action traffic, it adds an
\code{event_kind} and an \code{action_name} to the JSON event.  The event kind
identifies the action step rather than only the transport endpoint.  Goal
UUIDs are copied from requests, feedback, and status messages.  For a service
response, the monitor retains the UUID obtained from the corresponding
request.  A status message can contain several goals, so its event carries all
UUIDs reported in the message.  Table~\ref{tab:events} summarises the event
kinds produced by the implementation.

\begin{table}[H]
  \centering
  \caption{Action traffic represented in the event stream.}
  \label{tab:events}
  \small
  \begin{tabularx}{\linewidth}{@{}
    >{\raggedright\arraybackslash}p{0.24\linewidth}
    >{\raggedright\arraybackslash}p{0.32\linewidth}
    >{\raggedright\arraybackslash}X@{}}
    \hline\hline
    Traffic & Event kind & Correlation information \\
    \hline\hline

    \code{send_goal} request &
    \code{start_action_request} &
    action name and goal UUID \\
    \hline

    \code{send_goal} response &
    \code{start_action_response} &
    action name and UUID retained from the request \\
    \hline

    \code{get_result} request/response &
    \code{get_result_request} /\newline
    \code{get_result_response} &
    action name and correlated goal UUID \\
    \hline

    \code{cancel_goal} request/response &
    \code{cancel_action_request} /\newline
    \code{cancel_action_response} &
    action name and request selector; the response includes
    the accepted goal UUIDs \\
    \hline

    feedback topic &
    \code{feedback} &
    action name and goal UUID \\
    \hline

    status topic &
    \code{status} &
    action name and reported goal UUIDs \\
    \hline\hline
  \end{tabularx}
\end{table}

For example, an oracle may require a navigation goal to start only when the
robot arm is in a safe pose.  The \code{start_action_request} event contains
the navigation request, the action name, and its UUID.  If the request is
allowed, the server response is reported as a separate
\code{start_action_response} event.  Later feedback, status, result, and
cancellation events can be related to the same goal.
The monitor records an accepted goal only after two conditions hold: the
server has returned \code{accepted=true}, and the oracle has allowed the
corresponding response event.  Accepted goals are stored in a dictionary
keyed by the serialised UUID.  Each entry contains the action name, the event
time, flags that distinguish client- and monitor-initiated cancellation, and
flags for cancellation and completion.  This state is used when later topic
or service events require an intervention. An entry is removed after an allowed result response or when a successful
cancellation response lists the goal in \code{goals_canceling}.

The interface to the oracle is unchanged.  In offline mode, the enriched
events are written to the log.  In online mode, the monitor sends each event
as JSON through the WebSocket connection and waits for a verdict.  The oracle
can use any specification formalism that consumes this event stream; it does
not need to know how the five underlying endpoints were generated.

\subsection{Protocol-Aware Filtering}
\label{sec:monitor}

Filtering an action event cannot always mean dropping its underlying message.
Before a goal reaches the server, the monitor can reject it directly.  After
the server has accepted it, the monitor must instead request cancellation.
For other action steps, it must construct a response with the fields expected
by the corresponding \rostwo service.  Table~\ref{tab:enforcement} lists the
action-specific reactions implemented in filter mode.

\begin{table}[H]
  \centering
  \caption{Protocol-aware response to a violating oracle verdict
  in filter mode.}
  \label{tab:enforcement}
  \small
  \begin{tabularx}{\linewidth}{@{}l X@{}}
    \hline\hline
    Violation point & Monitor response \\
    \hline\hline

    \code{send_goal} request &
    Do not call the server and return a synthetic response
    with \code{accepted=false}. \\
    \hline

    Accepted \code{send_goal} response &
    Send \code{cancel_goal} for the UUID and return
    \code{accepted=false} to the client. \\
    \hline

    Feedback or status &
    Attempt to cancel the reported goal or goals and suppress
    publication of the violating topic message. \\
    \hline

    \code{get_result} request &
    Do not forward the request and return a synthetic response
    with \code{STATUS_UNKNOWN}. \\
    \hline

    \code{get_result} response &
    Return a synthetic response with \code{STATUS_ABORTED}. \\
    \hline

    \code{cancel_goal} request &
    Do not forward the request and return
    \code{ERROR_REJECTED}. \\
    \hline

    \code{cancel_goal} response &
    Repeat the server-side cancellation request and return
    \code{ERROR_REJECTED}. \\
    \hline\hline
  \end{tabularx}
\end{table}

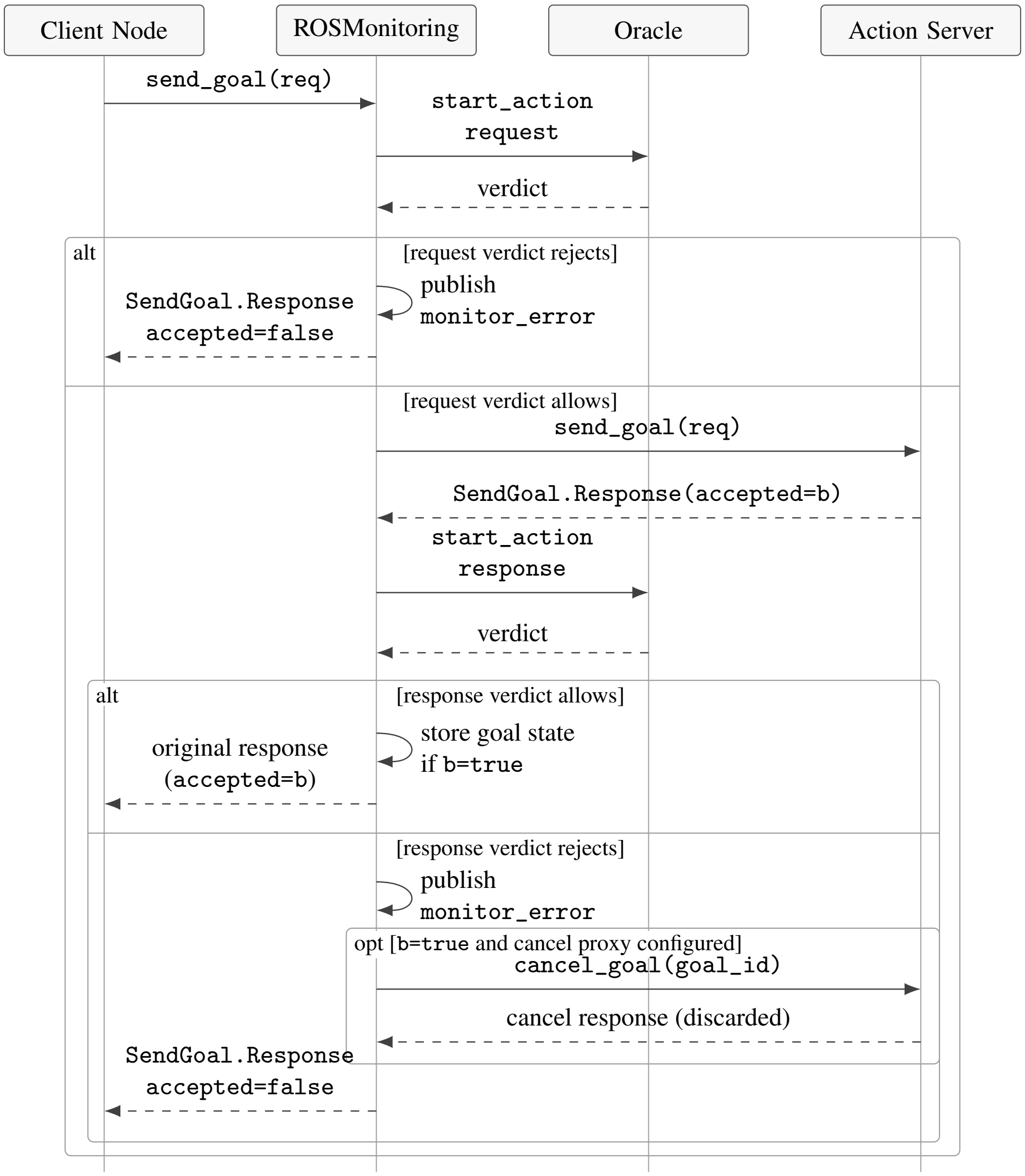
\begin{figure}[!htbp]
  \centering
  \scalebox{1.0}{

\begin{tikzpicture}[
  x=1cm,
  y=1cm,
  font=\footnotesize,
  participant/.style={
    draw=black!65,
    rounded corners=1.5pt,
    fill=black!3,
    minimum width=2.45cm,
    minimum height=0.62cm,
    text width=2.20cm,
    align=center
  },
  msg/.style={
    -{Latex[length=2mm]},
    draw=black!75,
    line width=0.45pt
  },
  reply/.style={
    -{Latex[length=2mm]},
    draw=black!75,
    dashed,
    line width=0.45pt
  },
  life/.style={
    draw=black!35,
    line width=0.4pt
  },
  frame/.style={
    draw=black!38,
    rounded corners=2pt,
    line width=0.4pt
  },
  tag/.style={
    fill=white,
    inner sep=1.2pt,
    font=\scriptsize
  }
]
  \node[participant] (client) at (0,0) {Client Node};
  \node[participant] (monitor) at (3.35,0) {\rosmon};
  \node[participant] (oracle) at (6.70,0) {Oracle};
  \node[participant] (server) at (10.05,0) {Action Server};

  \coordinate (clientend) at (0,-14.05);
  \coordinate (monitorend) at (3.35,-14.05);
  \coordinate (oracleend) at (6.70,-14.05);
  \coordinate (serverend) at (10.05,-14.05);

  \draw[life] (client.south) -- (clientend);
  \draw[life] (monitor.south) -- (monitorend);
  \draw[life] (oracle.south) -- (oracleend);
  \draw[life] (server.south) -- (serverend);

  \draw[msg] (0,-0.90) --
    node[above] {\code{send\_goal(req)}}
    (3.35,-0.90);

  \draw[msg] (3.35,-1.55) --
    node[above,align=center]
      {\code{start\_action}\\\code{request}}
    (6.70,-1.55);

  \draw[reply] (6.70,-2.18) --
    node[above] {verdict}
    (3.35,-2.18);

  \draw[frame] (-0.48,-2.55) rectangle (10.53,-13.85);
  \node[tag,anchor=north west] at (-0.43,-2.58) {alt};
  \node[tag,anchor=north] at (5.00,-2.58)
    {[request verdict rejects]};

  \draw[frame] (-0.48,-4.38) -- (10.53,-4.38);
  \node[tag,anchor=north] at (5.00,-4.41)
    {[request verdict allows]};

  \draw[msg] (3.35,-3.15)
    to[out=0,in=0,looseness=4.0]
    node[right,align=left]
      {publish\\\code{monitor\_error}}
    (3.35,-3.50);

  \draw[reply] (3.35,-4.02) --
    node[above,align=center]
      {\code{SendGoal.Response}\\\code{accepted=false}}
    (0,-4.02);

  \draw[msg] (3.35,-5.18) --
    node[above] {\code{send\_goal(req)}}
    (10.05,-5.18);

  \draw[reply] (10.05,-6.00) --
    node[above]
      {\code{SendGoal.Response(accepted=b)}}
    (3.35,-6.00);

  \draw[msg] (3.35,-6.92) --
    node[above,align=center]
      {\code{start\_action}\\\code{response}}
    (6.70,-6.92);

  \draw[reply] (6.70,-7.66) --
    node[above] {verdict}
    (3.35,-7.66);

  \draw[frame] (-0.20,-8.00) rectangle (10.28,-13.68);
  \node[tag,anchor=north west] at (-0.15,-8.03) {alt};
  \node[tag,anchor=north] at (5.00,-8.03)
    {[response verdict allows]};

  \draw[frame] (-0.20,-9.88) -- (10.28,-9.88);
  \node[tag,anchor=north] at (5.00,-9.91)
    {[response verdict rejects]};

  \draw[msg] (3.35,-8.65)
    to[out=0,in=0,looseness=4.0]
    node[right,align=left]
      {store goal state\\if \code{b=true}}
    (3.35,-9.00);

  \draw[reply] (3.35,-9.52) --
    node[above,align=center]
      {original response\\(\code{accepted=b})}
    (0,-9.52);

  \draw[msg] (3.35,-10.48)
    to[out=0,in=0,looseness=4.0]
    node[right,align=left]
      {publish\\\code{monitor\_error}}
    (3.35,-10.83);

  \draw[frame] (2.98,-11.05) rectangle (10.28,-12.72);
  \node[tag,anchor=north west] at (3.03,-11.08)
    {opt [\code{b=true} and cancel proxy configured]};

  \draw[msg] (3.35,-11.80) --
    node[above] {\code{cancel\_goal(goal\_id)}}
    (10.05,-11.80);

  \draw[reply] (10.05,-12.45) --
    node[above] {cancel response (discarded)}
    (3.35,-12.45);

  \draw[reply] (3.35,-13.30) --
    node[above,align=center]
      {\code{SendGoal.Response}\\\code{accepted=false}}
    (0,-13.30);
\end{tikzpicture}}
  \caption{Checking a \code{send_goal} interaction in filter mode.  Both the
  request and the server response are checked.  A server-accepted goal is
  cancelled if the response event violates the monitored property.}
  \label{fig:send-goal}
\end{figure}

Figure~\ref{fig:send-goal} shows the two checks performed while starting a
goal.  The first event is the \code{start_action_request}.  If its verdict is
violating, the monitor publishes a monitor error, does not call the action
server, and returns \code{accepted=false}.  Otherwise, the monitor forwards
the request and receives a \code{SendGoal.Response} from the server.  This
response is sent to the oracle as a \code{start_action_response}, regardless
of whether the server accepted or rejected the goal.

If the second verdict is non-violating, the original server response is
returned to the client.  When its \code{accepted} field is true, the goal is
also entered in the monitor state.  If the second verdict is violating and
the server had accepted the goal, the monitor calls the generated
\code{cancel_goal} proxy and waits for that service call to finish.  It then
returns a new \code{SendGoal.Response} with \code{accepted=false}. If the server had already rejected the goal, no cancellation is needed and the client simply receives a \code{SendGoal.Response} with \code{accepted=false}.

The table lists the points for which the extension adds an action-specific
reaction. For \code{get_result}, request and response violations are handled
differently. If a \code{get_result} request violates the property, the monitor
does not forward it to the server and returns a protocol-compliant
\code{GetResult.Response} with \code{STATUS_UNKNOWN}. The goal itself is not
interrupted. If the server response violates the property instead, the monitor
returns a \code{GetResult.Response} with \code{STATUS_ABORTED}.

The current online execution path is synchronous.  A topic callback
serialises and enriches the message, acquires the shared WebSocket lock, sends
the event, and waits for the oracle reply.  An intercepted service normally
performs this exchange twice: once for the request and once for the response,
with the downstream service call between them.  All endpoints generated for one monitor share the same connection and lock.
This preserves one order for the oracle exchanges and, under concurrent
workloads, can delay independent goals. Section~\ref{sec:evaluation} measures
the overall cost of this synchronous execution path under sequential goal
execution; it does not evaluate contention between concurrently active goals.

\section{Evaluation}
\label{sec:evaluation}

The evaluation addresses two complementary questions.  First, can the action
extension enforce safety properties that depend on the state of independent
\rostwo frameworks while allowing compliant action goals to proceed?  Second,
how does the runtime cost of the extension change with the number of action
calls, the amount of data produced by each action, and the number of
client--server pairs?

We answer the first question through a mobile-manipulation case study involving
Nav2 and MoveIt~2.  We answer the second through a synthetic benchmark in
which the workload can be varied systematically.  The two experiments
therefore serve different purposes: the case study evaluates the functional
behaviour of action-aware enforcement, while the benchmark measures its cost
and scalability.

In both experiments, the baseline configuration executes the same scenario
without \rosmon.  In the monitored configuration, the relevant action traffic
passes through \rosmon and is checked by the oracle.

\subsection{Qualitative Experiments: Cross-Framework Safety}
\label{sec:qualitative-experiments}

The qualitative experiments study whether \rosmon can enforce properties that
involve navigation and manipulation at the same time.  These properties cannot
be enforced by considering each framework in isolation because they depend on
their combined state.

\paragraph{Platform and scope.}
The case study uses the Clearpath simulator with a Husky A200 mobile base,
a Kinova Gen3 Lite arm, and a Kinova 2F Lite gripper%
~\cite{ClearpathManipulationSimulation}.  The system runs on
\rostwo Jazzy and Gazebo Harmonic.  Nav2 is used for navigation, MoveIt~2 for
arm motion planning, and \code{ros2_control} for low-level control.
All these components are standard \rostwo packages and their source code is not modified. The case-study repository contains the selected robot configuration in \code{robot.yaml}, the \rosmon configuration and properties, and a small set of nodes used to generate the experimental scenarios.  The robot-specific
\code{robot.yaml} file selects the required Clearpath configuration.
Clearpath then generates the corresponding robot description, controllers,
MoveIt configuration, and launch files.
Nav2 and MoveIt~2 manage navigation and manipulation independently.  However,
neither framework has complete knowledge of the global robot state.  Nav2 does
not determine whether the arm is in a safe configuration for navigation,
while MoveIt~2 does not determine whether an arm or gripper request is safe
while a navigation goal is active.  The case study uses \rosmon as an external
coordination layer that observes both frameworks and enforces constraints
involving their combined state.

\paragraph{Properties and experimental design.}
We consider three cross-framework safety properties, expressed as LTL
invariants.  P1 permits navigation only when the arm is in one of two
predefined configurations.  In the \code{STOWED} configuration, the arm is
folded close to the robot to minimise its footprint.  In the \code{CARRY}
configuration, the arm remains compact while keeping the gripper and any
transported object in a position that does not obstruct the lidar.  P2 rejects new arm-trajectory goals while a navigation goal is active. P3 rejects new gripper-opening goals while a navigation goal is active.
The properties are defined as follows:
\[
\begin{array}{ll}
P_1: &
\mathbf{G}\bigl(
\mathsf{nav\_request}
\Rightarrow
(\mathsf{arm\_stowed} \lor \mathsf{arm\_carry})
\bigr) \\
P_2: &
\mathbf{G}\bigl(
\mathsf{arm\_request}
\Rightarrow
\neg\mathsf{navigation\_active}
\bigr), \\
P_3: &
\mathbf{G}\bigl(
\mathsf{gripper\_open\_request}
\Rightarrow
\neg\mathsf{navigation\_active}
\bigr).
\end{array}
\]
The operator $\mathbf{G}$ is the LTL globally operator and requires its
argument to hold at every point of the execution.  The predicates
\code{nav_request}, \code{arm_request}, and
\code{gripper_open_request} identify requests for a new navigation goal, arm
trajectory, and gripper-opening operation, respectively.  These predicates
are true only for the corresponding request event.  The predicates
\code{arm_stowed} and \code{arm_carry} are derived from the observed arm-joint
values.  Finally, \code{navigation_active} indicates that Nav2 has accepted a
navigation goal that has not yet terminated.
Table~\ref{tab:case-study-properties} summarises how the three properties are
used in the experiments.
\begin{table}[H]
  \centering
  \caption{Properties and their roles in the qualitative experiments.}
  \label{tab:case-study-properties}
  \small
  \begin{tabularx}{\linewidth}{@{}l X X@{}}
    \hline\hline
    Property & Enforced condition & Role in the experiments \\
    \hline\hline

    P1 &
    Navigation is permitted only when the arm is in
    \code{STOWED} or \code{CARRY}. &
    The permitted case is exercised in both scenarios:
    the navigation goal is accepted and executed. \\
    \hline

    P2 &
    New arm-trajectory goals are rejected while a navigation
    goal is active. &
    Scenario~1 sends a new arm-trajectory goal during navigation
    and checks that it is rejected. \\
    \hline

    P3 &
    New gripper-opening goals are rejected while a navigation
    goal is active. &
    Scenario~2 sends a new gripper-opening goal during navigation
    and checks that it is rejected. \\
    \hline\hline
  \end{tabularx}
\end{table}
The experiments are not intended as an exhaustive test of every possible
property violation.  Instead, they consider two representative workflows.  In
both workflows, P1 defines a valid initial condition and allows navigation to
start normally.  Once navigation is active, each scenario introduces a
different conflicting request to exercise the enforcement of P2 or P3.

\paragraph{Monitoring implementation.}
The properties P1--P3 use Boolean values such as \code{nav_request}
and \code{arm_stowed}. Before checking these properties, the oracle
derives these values from the received JSON events. This step is
implemented by the \code{abstract\_message()} function.

The function recognises requests to navigate, move the arm, or open
the gripper. It uses joint positions to check whether the arm is in
\code{STOWED} or \code{CARRY}, and action responses and status updates
to track whether navigation is active.
The oracle is used to enforce the conjunction
$\Phi = P_1 \land P_2 \land P_3$.
Since all three properties are invariants whose violations can be determined
from the current request and the maintained robot state, \rosmon checks the
body of the relevant invariant before forwarding an action goal.  If
forwarding the goal would violate $\Phi$, \rosmon returns a rejected action
response before the goal reaches its action server.  In the baseline
configuration, the same action goals reach their corresponding servers
without this additional check.
The predicate \code{navigation_active} directly uses the action-lifecycle
support introduced by the extension.  It becomes true when Nav2 accepts a
navigation goal and remains true until Nav2 reports a terminal status.  The
monitor must therefore relate the goal-acceptance event to a later lifecycle
event rather than treating the underlying service and topic messages as
independent events.

\paragraph{Scenario 1: navigation and arm coordination.}
The first automated scenario is executed in both the baseline and monitored
configurations.  The arm initially moves to the predefined \code{STOWED}
configuration.  Nav2 then receives a navigation goal that requires the robot
to travel five metres.  Since the arm is in \code{STOWED}, the goal satisfies
P1 and navigation starts in both configurations.
While the navigation goal is active, the scenario asks MoveIt~2 to move the
arm from \code{STOWED} to \code{READY}.  In the baseline execution, MoveIt~2
sends the resulting trajectory to the \code{ros2_control} controller and the
arm moves while the robot is navigating.\footnote{Scenario 1, baseline:
\url{https://www.youtube.com/watch?v=JR_5EdMI9pI&list=PLVY6JOe5lQPo&index=1&pp=iAQBsAgC}.} In the monitored execution,
\rosmon rejects the corresponding trajectory action goal before it reaches
the controller, as required by P2.\footnote{Scenario 1, monitored:
\url{https://www.youtube.com/watch?v=vEaGh52vFp4&list=PLVY6JOe5lQPo&index=2&pp=iAQBsAgC}.}  The requested arm movement is therefore
not executed.
Rejecting the arm trajectory does not interrupt the independent navigation
goal.  In both executions, the robot continues towards the same destination.
After navigation terminates, \code{navigation_active} becomes false and a new
request to move the arm to \code{READY} is permitted.

\paragraph{Scenario 2: box transport.}
The second automated scenario is also executed in the baseline and monitored
configurations.  The robot first picks up a box from the floor and moves the
arm to the predefined \code{CARRY} configuration.  Nav2 then receives a
five-metre navigation goal.  Since the arm is in \code{CARRY}, the goal
satisfies P1 and is accepted in both configurations.
After the robot has travelled two metres, the scenario sends a request to open
the gripper.  In the baseline execution, the request is accepted and the box
is released.\footnote{Scenario 2, baseline: \url{https://www.youtube.com/watch?v=NRLHzBWI9KY&list=PLVY6JOe5lQPo&index=3&pp=iAQBsAgC}.}
In the monitored execution, \rosmon rejects the same gripper
action goal because \code{navigation_active} is true, as required by P3.\footnote{Scenario 2, monitored:
\url{https://www.youtube.com/watch?v=2HhuFJ9jb4I&list=PLVY6JOe5lQPo&index=4&pp=iAQBsAgC}.}  The
gripper therefore remains closed while the robot continues towards the same
destination.

\paragraph{Qualitative findings.}
The two scenarios exercise both the transparent and enforcing behaviours of
the extension.  Navigation goals satisfying P1 are forwarded and executed
normally.  Once navigation is active, the monitor uses the state of the
accepted goal to reject an incompatible arm trajectory in Scenario~1 and an
incompatible gripper goal in Scenario~2.
The experiments also show that rejecting one action goal does not stop an
independent action that is already running.  In both scenarios, the unsafe
manipulation request is rejected while the navigation goal continues.
Moreover, \rosmon returns a protocol-compliant goal rejection rather than
simply dropping an underlying service request.  This behaviour requires the
action-level information introduced by the extension.
The case study provides functional evidence that \rosmon can coordinate
actions managed by independent \rostwo frameworks, preserve permitted
behaviour, and reject representative cross-framework conflicts. These
constraints are introduced externally, without modifying Nav2, MoveIt~2,
\code{ros2_control}, or the robot controllers.

A dedicated controller could also enforce P1--P3. Here, users define
the properties, while ROSMonitoring provides the action interception
and the mechanisms needed to reject conflicting goals.

\FloatBarrier

\subsection{Quantitative Experiments: Scalability}
\label{sec:quantitative-experiments}
\label{sec:scalability}

The qualitative experiments demonstrate the extension in a robotic scenario.
To study monitoring cost under controlled workloads, we use the \rostwo
Fibonacci action interface with a benchmark server.

Each goal specifies a sequence length $s$. Starting from 0 and 1, the server
computes the first $s$ Fibonacci numbers, publishes the growing sequence as
feedback after each additional element, and returns the completed sequence
as the result. No artificial execution delay is introduced. Varying $s$
therefore changes both the result size and the feedback traffic, while
varying the number of goals changes how often the action lifecycle is
executed. This simple workload allows us to study monitoring overhead
without the additional behaviour of navigation or manipulation software.

\paragraph{Experimental design.}
The experiments were run on a machine equipped with an AMD Ryzen 5 3600
processor, an NVIDIA GeForce RTX 2080 graphics card, and 16GB of DDR4 memory
clocked at 3200MHz.

The monitored configuration uses an always-true oracle. It parses each
incoming JSON event and returns a non-violating verdict without evaluating
a temporal property. The benchmark therefore measures the monitoring
infrastructure with minimal verdict logic, including interception,
serialisation, event exchange, and forwarding. Application-specific
property evaluation may introduce additional cost.
The benchmark creates $n$ distinct action client--server pairs.
All $n$ ActionClients are hosted by one client node, while all $n$
ActionServers are hosted by one server node. Goals are executed sequentially:
the client waits for each result before sending the next goal, including when
switching between action pairs. Therefore, $n$ varies the number of configured
action interfaces, not the number of simultaneously active goals. Each pair
receives $g$ action goals, so one measured execution contains
$a = n \cdot g$ goals. Each server returns a Fibonacci sequence of length $s$;
throughout the paper, we call it sequence length because it is the number of
returned elements.
We evaluate every combination
$n\in{1,\ldots,20}$,
$g\in{1,\ldots,20}$, and
$s\in{2,\ldots,20}$.  This gives 7,600 conditions. Each condition is repeated five times, producing
38,000 paired repetitions and 76,000 individual executions across the
baseline and monitored configurations. Across all repetitions, the baseline executes 4,189,500 goals and the
monitored configuration executes the same number, for 8,379,000 goal
executions in total.
The baseline timings contain occasional large positive spikes.  We therefore
use the median of the five repetitions for each condition as the main
estimate.  All figures and summary values below use these condition medians.
We report the added time
$\Delta T=T_M-T_0$ and the slowdown $T_M/T_0$, where $T_0$ is the baseline
elapsed time and $T_M$ is the monitored elapsed time.

Timing starts after all action servers are available and ends after the
client receives the final goal result. Process launch and the initial wait
for server availability are outside the measured interval.

\begin{figure}[!htbp]
\centering
\includegraphics[width=\linewidth]{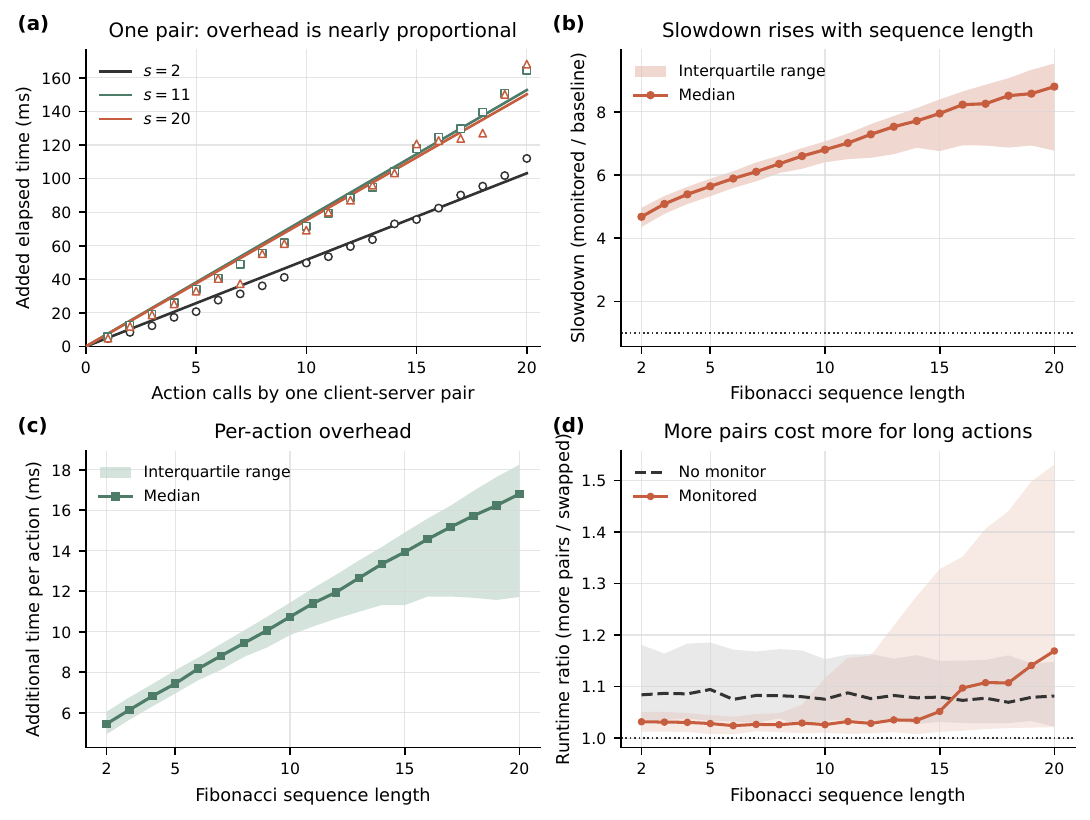}
\caption{Benchmark results based on the median of five complete executions
for each condition.  Panel~(a) shows repeated action calls from one
client--server pair.  Panels~(b) and~(c) show how slowdown and added time per
action change with sequence length.  Panel~(d) compares configurations with
more pairs against configurations with more goals per pair while keeping
the total goal count fixed.  Shaded regions show the interquartile range.}
\label{fig:benchmark}
\end{figure}

Figure~\ref{fig:benchmark} presents the results.  Its four panels address
three questions: whether overhead accumulates with repeated action calls,
whether it changes with sequence length, and whether distributing the same
number of goals across more client--server pairs introduces an additional
cost.

\paragraph{Overall cost.}
Across the 7,600 conditions, the median added time is 0.885s and the median
slowdown is 6.52, with an interquartile range from 5.58 to 7.74.  Monitoring
is slower in 7,599 conditions.  The single exception contains large positive
spikes in the baseline repetitions and does not indicate a performance
advantage.
At the largest tested setting, $(n,g,s)=(20,20,20)$, one execution contains
400 goals.  Median elapsed time rises from 0.877s without monitoring to
8.766s with monitoring.  This corresponds to 7.890s of added time and a
slowdown of 10.00.
These measurements show that the cost of the current implementation is not
negligible under a dense workload of short actions.  The slowdown values must
also be read in the context of this benchmark.  Fibonacci goals complete
quickly, and the fixed work required to intercept, serialise, check, and
forward their events forms a large part of their execution time.  The ratios
should therefore not be transferred directly to longer robotic actions.
Nevertheless, they show that the current synchronous monitoring path requires
optimisation before it can be used for workloads containing many short or
frequent actions.

\paragraph{Cost of repeated action calls.}
The setting $n=1$ isolates one client--server pair.  Holding $s$ fixed and
raising $g$ from 1 to 20 tests how the total monitoring overhead changes when
the same pair calls its action repeatedly.  As shown in
Figure~\ref{fig:benchmark}(a), the total added time grows almost linearly with
$g$.  For each of the 19 sequence lengths, a straight line explains between
98.5\% and 99.8\% of the observed variation.  Lines constrained to pass
through the origin also match the measurements closely, supporting an
approximately proportional relationship.
The result can be summarised as
$\Delta T(1,g,s)\approx g\cdot c_s$,
where $c_s$ is the marginal cost of one action call for a fixed sequence
length and topology.  For the sequence lengths shown in the figure, this cost
is approximately 5--8ms per action.  Thus, each additional action
contributes approximately the same amount of overhead.  The result does not
indicate that the monitor becomes slower over time or that later calls are
more expensive.
This behaviour follows from the event-based operation of the monitor.  Each
additional goal produces new protocol traffic and new oracle interactions.
Their processing costs are added to the elapsed time of the complete
execution.  The measurements therefore do not support interpreting the
overhead mainly as a one-time setup cost associated with creating the
client--server pair.
The coefficient $c_s$ measures the added time associated with one complete
action lifecycle.  It combines the cost of checking the goal request and
response, feedback and status messages, and the result request and response.
It is therefore a per-\emph{action} cost rather than a per-\emph{event} cost.

\paragraph{Effect of sequence length.}
Panels~(b) and~(c) of Figure~\ref{fig:benchmark} show that action count alone
does not explain the overhead.  At $s=2$, the median slowdown is 4.68 and the
median added time per action is 5.45ms.  At $s=20$, these values increase to
8.80 and 16.80ms, respectively.  Over the same range, median monitored time
increases by a factor of 2.53, while baseline time increases by a factor of
1.39.
This difference is consistent with how online monitoring is implemented.
Each intercepted message is converted to JSON and sent to the oracle, and the
monitor waits for the oracle response before continuing.  Longer sequences
produce larger feedback and result payloads.  They can also produce more
feedback events because the server publishes partial sequences while
constructing the result.  Both effects increase the work performed by the
monitor.
The measurements do not separate the cost of larger payloads from the cost of
additional feedback events.  They nevertheless show that there is no
universal per-action overhead.  The cost depends on the action traffic exposed
to the monitor, and therefore on both the number and the size of the events
generated by an action.  This result points to compact event representations
and selective monitoring of relevant lifecycle events as possible
optimisations.

\paragraph{Effect of topology.}
The total number of goals $a=n\cdot g$ can be kept fixed while changing how
those goals are distributed. For every setting with $n>g$, we compare
$(n,g,s)$ with $(g,n,s)$. The first configuration uses more configured
client--server pairs, while the second sends more goals through each pair.
Both execute the same number of goals and use the same sequence length.
This gives 190 matched comparisons for each sequence length and 3,610 matched
comparisons in total.
Of these, a contiguous block affected by anomalous timing measurements was
excluded from the analysis. Specifically, we excluded matched comparisons
involving $(n,g)=(8,19)$, $(8,20)$, or $(9,g)$ with
$g\in\{1,\ldots,17\}$. This leaves 3,268 matched comparisons.
Figure~\ref{fig:benchmark}(d) shows that the two configurations do not have the
same execution time.  In the monitored configuration, the median ratio
between the run with more pairs and its swapped run is 1.032 at $s=2$ and
1.169 at $s=20$.  Thus, at the longer sequence length, distributing the same
goals across more pairs is 16.9\% slower at the median.  In the baseline
configuration, the corresponding ratio remains close to 1.08 across the
sequence lengths.
The number of goals is therefore the main scaling factor, but topology also
matters. More pairs introduce additional action endpoints, middleware entities,
discovery traffic, and scheduling work, while the number of nodes and processes
remains fixed. Oracle exchanges share one WebSocket lock and are therefore
serialised. However, since goals are issued sequentially, this experiment does
not measure contention between concurrently active goals and does not isolate
the causes of the observed topology effect. The shared connection remains an
important target for investigation under concurrent workloads.

\paragraph{Discussion and limitations.}
The two evaluations provide complementary results. The case study demonstrates
protocol-aware enforcement, while the benchmark characterises its runtime
cost. At fixed sequence length, added time grows approximately with the number
of goals. Sequence length and the distribution of goals across configured
pairs are associated with smaller additional effects. These results concern
the current sequential and synchronous implementation; they do not represent
an inherent lower bound for action monitoring. Future optimisation should
reduce event volume and the cost of synchronous exchanges with the oracle.

The benchmark measures the total added elapsed time. It does not
separate communication between the monitor and the oracle from
monitor processing and oracle evaluation. Further profiling is
needed to measure these contributions and identify the main
sources of overhead.

\section{Related Work}
\label{sec:related-work}

Runtime verification has been integrated with ROS at several abstraction
levels. An early ROS-specific approach is ROSRV, introduced by Huang
et al.~\cite{DBLP:conf/rv/HuangEZMLSR14}. ROSRV provides transparent
instrumentation that intercepts ROS commands and messages and checks them
against formally specified safety and security properties using generated
monitors. \rosmon subsequently adopted a different architecture, separating
ROS instrumentation from the oracle that implements the monitoring
formalism~\cite{FerrandoEtAl2020}. This makes the ROS-facing part of the
framework independent of the property language used by the oracle.
\rosmon 2.0 extended the same approach to \rostwo, including support for
service monitoring~\cite{GhaffariSaadatEtAl2024}. These works are
the closest predecessors of the present paper. Our extension retains the
topic and service instrumentation of \rosmon 2.0, but raises the monitored
abstraction from individual endpoints to a goal-correlated action lifecycle.

Other work has connected ROS and \rostwo to particular specification and
monitoring technologies. Perez et al. generate \rostwo monitoring nodes
from structured requirements by integrating FRET, Ogma, and
Copilot~\cite{DBLP:journals/corr/abs-2209-14030}. Stadler et al. proposed
flexible runtime-monitoring support for ROS
applications~\cite{DBLP:conf/rose-ws/StadlerVC22} and subsequently presented
ROMoSu, a framework for configurable runtime monitoring of ROS-based
systems~\cite{DBLP:conf/rose-ws/StadlerV23}. Begemann et al. introduced the
TeSSLa-ROS-Bridge, which connects ROS-based robotic systems to
TeSSLa~\cite{DBLP:conf/ictac/BegemannKLS23}. RTAMT provides quantitative
monitoring of Signal Temporal Logic properties and includes a ROS
integration~\cite{DBLP:journals/sttt/YamaguchiHN24}. More recently, Baumeister
et al. presented an adapter connecting RTLola monitors to ROS topics and
services~\cite{DBLP:conf/rv/BaumeisterFJKST25}. These integrations focus on
monitoring nodes, streams, topics, or services; to the best of our knowledge,
they do not represent a \rostwo action as a goal-correlated lifecycle.

Runtime monitoring also appears in broader runtime-assurance architectures
for robotics. Desai et al. combine model checking of high-level robotic
software with online monitoring of assumptions about lower-level
controllers~\cite{DBLP:conf/rv/DesaiDS17}. SOTER places a language-based
runtime-assurance architecture on top of ROS to protect distributed mobile
robotic systems~\cite{DBLP:conf/rv/ShivakumarTDS20}. For robot swarms, Hu et al.
study hierarchical runtime verification~\cite{DBLP:journals/tr/HuDYSZ20} and
decentralised enforcement~\cite{DBLP:journals/jzusc/HuDYSD20}. These approaches
enforce safety at the level of controllers, behaviours, or swarm properties.
In contrast, our enforcement operates on the \rostwo communication protocol:
a violation may reject a goal, cancel an accepted goal, suppress feedback or
status traffic, or produce a valid protocol response.

Recent work also considers the wider engineering context in which runtime
verification is deployed. Caldas et al. study runtime verification and field
testing for ROS-based robotic systems and identify guidelines for designing
applications that support runtime observation and
testing~\cite{DBLP:journals/tse/CaldasGSPRB24}. At the architectural level,
Safe-ROS separates normal autonomous control from independently verifiable
Safety Instrumented Functions~\cite{DBLP:journals/corr/abs-2511-14433}. These
approaches concern the wider runtime-assurance architecture, whereas our work
exposes and mediates the semantics of a specific \rostwo communication
abstraction.

To the best of our knowledge, existing ROS runtime-verification approaches do
not treat a \rostwo action as a single monitored protocol correlated by goal
identifier with action-specific enforcement. Our contribution is an
action-aware instrumentation and enforcement layer extending \rosmon's
formalism-agnostic architecture to this previously unsupported abstraction.

\section{Conclusion and Future Work}
\label{sec:conclusion}

We extended \rosmon with support for \rostwo actions. A single action
declaration generates the five protocol endpoints, while the monitor adds
lifecycle information, goal correlation, and protocol-aware enforcement.
The mobile-manipulation case study showed that permitted goals proceed
normally and that new conflicting arm and gripper goals can be rejected
without interrupting navigation or modifying the underlying frameworks.
The benchmark found approximately proportional growth in added time with the
number of goals at fixed sequence length, together with additional effects
associated with sequence length and goal distribution.

Future work will investigate a C++ implementation of the ROS-facing
monitor using \code{rclcpp} and a native oracle integrated with the
monitor. We will compare this approach with the existing external
oracle to assess the trade-off between performance and flexibility.
Further experiments will evaluate concurrent goals, cancellation,
late feedback, and server failures.

We also plan to investigate predictive runtime verification.
An action may satisfy the properties when it starts but lead to
a violation during execution or when its goal is reached.
Predicting such violations requires information about how the
action and its environment may evolve.

Another direction is to combine action monitoring with state
machines describing the robot's current condition. For example,
after a grasp action has finished, the robot may still be holding
a fixed handle. Starting navigation in that state could be unsafe
even though no arm action is running.

\section*{AI Usage}

OpenAI’s ChatGPT (GPT-5.6 SOL) was used to support the preparation and language editing
of this paper.  The authors remain responsible for checking all technical
claims, experimental results, figures, and references before submission.

\bibliographystyle{eptcs}
\bibliography{main}

\end{document}